\documentclass{article}
\usepackage{ijcai23}

\usepackage{times}
\usepackage{soul}
\usepackage{url}
\usepackage[hidelinks]{hyperref}
\usepackage[utf8]{inputenc}
\usepackage[small]{caption}
\usepackage{graphicx}
\usepackage{amsmath}
\usepackage{amsthm}
\usepackage{booktabs}
\usepackage{algorithm}
\usepackage{algorithmic}
\usepackage[switch]{lineno}
\usepackage{multirow}
\usepackage{amssymb}

\newtheorem{definition}{Definition}

\title{The Risk of Federated Learning to Skew Fine-Tuning Features and Underperform Out-of-Distribution Robustness}

\author{
Mengyao Du$^1$
\and
Miao Zhang$^1$\and
Yuwen Pu$^{2}$\And
Kai Xu$^1$
Shouling Ji$^2$
Quanjun Yin$^1$
\affiliations
$^1$National University of Defense Technology\\
$^2$Zhejiang university
\emails
\{dumengyao, zhangmiao15, xukai09\}@nudt.edu.com,
\{yw.pu, sji\}@zju.edu.cn,
yin\_quanjun@163.com
}

\begin{document}

\maketitle

\begin{abstract}

To tackle the scarcity and privacy issues associated with domain-specific datasets, the integration of federated learning in conjunction with fine-tuning has emerged as a practical solution. However, our findings reveal that federated learning has the risk of skewing fine-tuning features and compromising the out-of-distribution robustness of the model. By introducing three robustness indicators and conducting experiments across diverse robust datasets, we elucidate these phenomena by scrutinizing the diversity, transferability, and deviation within the model feature space. To mitigate the negative impact of federated learning on model robustness, we introduce GNP, a \underline{G}eneral \underline{N}oisy \underline{P}rojection-based robust algorithm, ensuring no deterioration of accuracy on the target distribution. Specifically, the key strategy for enhancing model robustness entails the transfer of robustness from the pre-trained model to the fine-tuned model, coupled with adding a small amount of Gaussian noise to augment the representative capacity of the model. Comprehensive experimental results demonstrate that our approach markedly enhances the robustness across diverse scenarios, encompassing various parameter-efficient fine-tuning methods and confronting different levels of data heterogeneity.

\end{abstract}

\section{Introduction}

\begin{figure}
  \centering
  \includegraphics[width=0.48\textwidth]{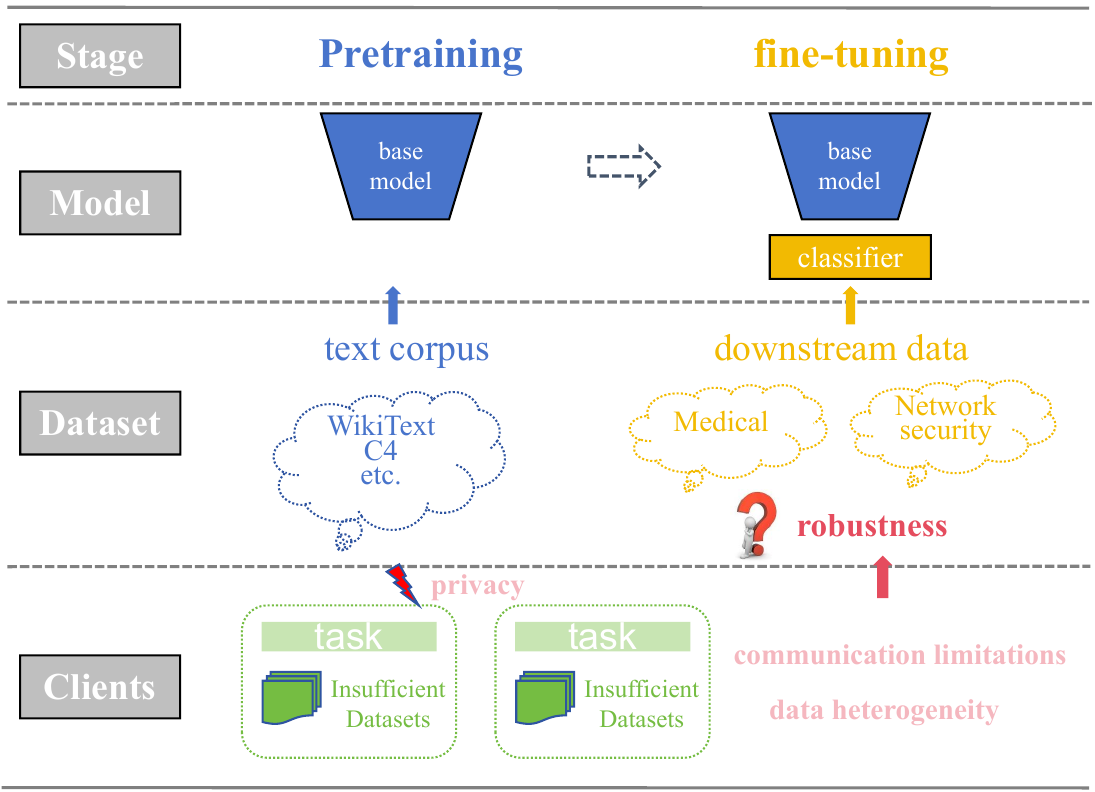}
  \caption{Two-Stage deployment process of large language models.}
  \label{fig:overall}
\end{figure}

Large language models have garnered considerable attention from both academic and industrial communities, providing artificial intelligence with the capability to adeptly integrate into myriad downstream applications. The deployment of large language models typically involves two discernible phases. As depicted in Figure~\ref{fig:overall}, the initial stage focuses on pre-training the model on an extensive corpus, facilitating the acquisition of broad language knowledge and structures. {\em Pre-trained Language Models} (PLMs), exemplified by the transformer architecture, have witnessed widespread open-sourcing, with notable instances such as Google's BERT~\cite{devlin2018bert}, the T5 model~\cite{t5raffel2020exploring}, and Meta's LLaMA model~\cite{touvron2023llama}. Following the pre-training phase, clients engage in fine-tuning the model using domain-specific datasets, thereby enhancing its adaptability to the nuanced data distribution. 

However, in fields such as network security and healthcare, proprietary domain datasets held by individual clients or enterprises typically demonstrate scarcity and involve privacy-sensitive information. Using the sentence classification task as an example, the volume of the fine-tuning dataset usually ranges from 30k to 400k instances. Consequently, the data held by an individual client is often insufficient to meet the criteria of data quality and quantity required for fine-tuning. To tackle the challenges posed by privacy constraints and data scarcity, recent studies have proposed {\em federated fine-tuning}, employing federated learning in the fine-tuning stage. Nevertheless, the impact of federated fine-tuning on the model's out-of-distribution robustness has not yet been thoroughly investigated. Intuitively, machine learning models must demonstrate generalization to unforeseen data existing in diverse distributions yet pertinent to the same task. This necessity becomes particularly critical in cases where the model robustness of new features and distributions plays a pivotal role in ensuring the reliable deployment of machine learning models with confidence.

Given the robustness requirements and constraints imposed by data scarcity, we contemplate an interesting question: {\em Does the federated fine-tuning impact the out-of-distribution robustness of the model?}

In this paper, we systematically introduce three robustness indicators—{\em Singular Value Entropy} (SVE), {\em Largest Singular Value Ratio} (LSVR), and {\em Gradient Deviation Angle} (GDA). By monitoring the variations in these three indicators at the classifier layer of the model, we capture insights into the diversity, transferability, and deviation in the model's feature space. Simultaneously, extensive experiments on different robust datasets are orchestrated. Our experimental results indicate that federated fine-tuning can skew fine-tuning features and underperform the model's robustness.

Specifically, in federated learning scenarios, two prominent features include data heterogeneity and communication limitations arising from geographically dispersed clients. Taking full fine-tuning as an example, as the heterogeneity of client data increases, the model's accuracy on in-distribution (ID) datasets exhibits controllable fluctuations. Conversely, the accuracy of the robust dataset decreases relatively swiftly. For instance, with a label distribution shift, the accuracy on the ID dataset increases from 89.6\% to 90.8\%, while on the robust dataset Stanford Sentiment Treebank (SST), it decreases from 72.3\% to 60.0\%. Moreover, during the fine-tuning phase, a significant exchange of model parameters occurs between clients and servers, often reaching scales of millions or even billions. This underscores the critical need to implement parameter-efficient fine-tuning (PEFT) methods practically. Specifically, these methods involve freezing the majority of parameters and updating only a small fraction of a large model. Representative approaches include BitFit~\cite{zaken-et-al:scheme}, Prefix tuning~\cite{li-et-al:scheme}, Adapter tuning~\cite{houlsby-et-al:scheme}, and LoRA~\cite{hu-et-al:scheme}. However, in our experiments, we observed that when the dataset exhibits a significant degree of heterogeneity, the model's robustness demonstrates an inverse correlation with the communication efficiency of PEFT methods. For example, Prefix tuning achieves a communication reduction of 12x but achieves only 37.4\% accuracy on the robust dataset SST. In contrast, BitFit achieves a communication reduction of 190x with an accuracy of 60.2\%. Building upon these findings, there is an urgent need for research to delve into the impact of federated learning on model robustness.

To improve model robustness, adaptable to various PEFT methods under increased data heterogeneity, we present GNP, a \underline{G}eneral \underline{N}oisy \underline{P}rojection-based robust algorithm. Our design involves the ingenious creation of a robust vector to transfer robustness from the pre-trained model to the fine-tuned model. Furthermore, we introduce carefully designed quantitative Gaussian noise to enhance the model's robustness without compromising performance. The versatility and adaptability of our approach have been validated through numerous experiments.

Our primary contributions can be encapsulated in three key aspects: (1) We introduced three robustness indicators and conducted extensive experiments on robust datasets, providing insights into model feature space and out-of-distribution robustness; (2) Our findings first reveal that the federated fine-tuning may distort model features and undermine the model out-of-distribution robustness; (3) We introduced a general robust algorithm, GNP, capable of substantially alleviating the impact of federated learning on model robustness.

\section{Related Works}

We review related work from three perspectives, namely, out-of-distribution robustness, federated learning, and parameter-efficient fine-tuning methods.

\subsection{Out-of-distribution Robustness}

The assessment of out-of-distribution (OOD) robustness currently lacks a conclusive determination. Prevailing concepts of OOD robustness more closely correspond to domain generalization~\cite{Muandet-et-al:scheme,zhou-et-al:scheme} or OOD generalization~\cite{ye-et-al:scheme}. These concepts indicate the model's capability to generalize to unforeseen distributional changes within the same task domain.

Recent research has unveiled advancements in OOD robustness for pre-trained language models, particularly those characterized by Transformer architecture, in comparison to their predecessors~\cite{brown-et-al:scheme,wang-et-al:scheme}. However, certain studies suggested that even when pre-trained models exhibit robustness, this resilience encounters challenges when smoothly transitioning to fine-tuned downstream tasks~\cite{chen-et-al:scheme,kumar-et-al:scheme}.

The endeavor to enhance model robustness is currently a subject of extensive research. Representative approaches include data augmentation, wherein existing datasets are expanded through algorithms such as affine transformations, reversals, and cropping, as exemplified in referencess~\cite{rebuffi-et-al:scheme,croce-et-al:scheme}. Additionally, some studies explored strategies for augmenting either the sizes~\cite{ji-et-al:scheme} or diversity~\cite{zhou-et-al:scheme1} of datasets to improve model performance. However, these methods predominantly operate at the data level. They may not be suitable for scenarios in federated learning, where data is distributed among clients and the central server lacks control over user data.

Furthermore, another avenue explored during the model training phase is the weight-space ensemble, which represents a distinct research paradigm. \cite{wortsman-et-al:scheme} proposed a method that involves linear interpolation between the original model and fine-tuned models to preserve salient features better. \cite{cai-et-al:scheme} investigated the relationship between model robustness and weights, introducing the Robust Weight Signature, a clever amalgamation of model weight interpolation and arithmetic.

Currently, the work closest to ours is described in \cite{cai-et-al:scheme}, albeit focusing solely on weight updates under adversarial sample perturbations. In this paper, our objective is to devise a general robust algorithm without prior knowledge about data heterogeneity and fine-tuning methods.

\subsection{Federated Learning}

Federated learning is a popular distributed learning paradigm that enables geographically dispersed clients to collaboratively train models while preserving privacy. A key challenge in federated learning scenarios is data heterogeneity, determined by diverse users and variations among them.

To alleviate the impact of data heterogeneity, numerous studies have dedicated efforts to investigate its effects. For instance, \cite{shi-et-al:scheme} observed that data heterogenization leads to a dimensional collapse in representations, significantly constraining the model's expressive capacity and affecting overall model performance. \cite{luo-et-al:scheme} discovered reduced feature similarity among different clients at the classifier layer. \cite{wang2022does} found that decentralized self-supervised learning has the potential to mitigate the impact of data heterogeneity.

While previous approaches offer valuable insights, the approach presented in this paper distinguishes itself. To the best of our knowledge, a substantial gap is evident in the existing literature concerning the impact of federated fine-tuning on the out-of-distribution robustness.

\subsection{Parameter-Efficient Fine-tuning Methods}

Building upon the categorization proposed by \cite{ding-et-al:scheme}, we systematically categorize PEFT methods into three groups: specification-based methods (BitFit), addition-based methods (Prefix tuning and Adapter tuning), and reparameterization-based methods (LoRA).

\begin{itemize}
    \item {\em BitFit}~\cite{zaken-et-al:scheme}: It freezes the majority of transformer encoder parameters and exclusively updates the bias parameters. This method proficiently capitalizes on the knowledge encapsulated in pre-trained models while mitigating the hazard of overfitting.
    \item {\em Prefix tuning}~\cite{li-et-al:scheme}: It involves constructing a segment of task-related virtual tokens before the input tokens. During training, only the parameters of the prefix modules are updated. This method shares similarities with prompt-based approaches.
    \item {\em Adapter tuning}~\cite{houlsby-et-al:scheme}: Integrating adapter modules within transformer layers provides a viable method for enabling transfer learning. This process preserves essential informational structures within the pre-trained model. 
    \item {\em LoRA}~\cite{hu-et-al:scheme}: The core concept of LoRA involves decomposing the matrix $\mathcal{W} \in \mathcal{R}^{d \times k}$ into the product of B and A, where B clutching a considerably low rank $r\ll min(d,k)$. This furnishes the prospect of refining immense models utilizing a remarkably minimal parameter quota.
\end{itemize}

\section{Understanding the Impact of Federated Learning on Out-of-distribution Robustness}

We elucidate the impact of data heterogeneity and PEFT methods on model robustness by designing robust indicators and conducting experiments on robust datasets.

\subsection{Designing Robust Indicators}
\label{section3.1 robust indicator}

The feature space $\mathbf{F} \in \mathbb{R}^{M \times D}$ belonging to the model's classifier layer undergoes Singular Value Decomposition (SVD): $\mathbf{F}=\mathbf{U\Sigma}\mathbf{V}^{\top}$. Consequently, $\mathbf{\Sigma}$ is obtained as a diagonal singular value matrix $\{\sigma_1,...,\sigma_D\}$, where $\sigma_1$ represents the maximum eigenvalue. In this investigation, three robust indicators derived from $\mathbf{F}$ are introduced to represent the diversity, transferability, and deviation within the feature space.



\begin{definition}[Singular Value Entropy, SVE]
    In multidimensional space complexity, Singular Value Entropy is introduced as an intrinsic metric to serve as a quantifier for measuring the complexity of feature space.
\end{definition}

\begin{equation}
    \label{eq: classifier SVE}
    \begin{aligned}\mathrm{SVE}&=-\sum_{i=1}^D\frac{\sigma_i}{\sum_{j=1}^D\sigma_j}\log\frac{\sigma_i}{\sum_{j=1}^D\sigma_j}\end{aligned}
\end{equation}

Typically, a higher SVE score suggests that the model encompasses a broader spectrum of dimensions and captures a greater diversity of data structures~\cite{chen-et-al:scheme}.

\begin{definition}[Largest Singular Value Ratio, LSVR]
    The LSVR is formally defined as the proportion of the maximum singular value to all singular values, serving as a metric for the model's transferability.
\end{definition}

\begin{equation}
    \label{eq: classifier LSVR}
    \mathrm{LSVR}=\frac{\sigma_{1}}{\sum_{i=1}^{D}\sigma_{i}}\
\end{equation}

Typically, a smaller LSVR indicates a more extensive distribution of the matrix across multiple dimensions, showcasing greater model transferability~\cite{chen2019transferability}.

\begin{definition}[Gradient Deviation Angle, GDA]
   GDA is defined as the ratio of the residual vector obtained by projecting the fine-tuned model onto the pre-trained model to the fine-tuned model itself. This ratio is employed to reflect the deviation during model training.
\end{definition}

\begin{equation}
    \mathrm{GDA} = \frac{\theta_{ft} - P_{\theta_{pre}}(\theta_{ft})}{\theta_{ft}}
\end{equation}

A larger GDA value reflects a greater deviation between the fine-tuned and pre-trained model. As the Transformer structure undergoes training on a large corpus, the pre-trained model demonstrates excellent robustness~\cite{qu2022rethinking}. We designed this indicator with the intuition of inheriting robustness from the pre-trained model. 

\subsection{Experiment Design}
\subsubsection{ID Dataset and Robust Datasets}

Sentiment analysis serves as a classic task in machine learning, encompassing the identification and extraction of sentiment information from textual data. The results primarily pertain to the emotional undertones of the text, categorized explicitly into positive, negative, or neutral sentiments. To address concerns regarding OOD robustness, we have chosen four datasets with the same task but different distributions, in line with the approach of \cite{yuan2023revisiting}: Amazon~\cite{amazon}, DynaSent~\cite{potts2020dynasent}, SemEval~\cite{semeval}, and SST~\cite{sst}. Specifically, our model undergoes training on the in-distribution (ID) dataset, Amazon, while the remaining three test datasets are denoted as robust datasets. The detailed comparison of the datasets can be found in Table \ref{tab:dataset}, where SimCSE scores for both the ID dataset and robust datasets are based on the study conducted by \cite{yuan2023revisiting}. Elevated SimCSE scores indicate an enhanced level of semantic similarity within the respective datasets.

\begin{table}
    \centering
    \scalebox{0.9}{\begin{tabular}{llrrrr}
        \toprule
        \multirow{2}{*}{Dataset} & \multirow{2}{*}{Source} & \multirow{2}{*}{Classes} & \multicolumn{2}{c}{Samples} & \multirow{2}{*}{SimCSE} \\  &        &          & Train         & Test     &     \\
        \midrule
        Amazon     & Product      & 3     & 30,000    & 38,905     & 100   \\
        Dynasent   & Adversarial  & 3     & 93,553    & 4,320      & 57.3  \\
        SemEval    & Twitter      & 3     & 6,000     & 20,622     & 24.74  \\
        SST        & Movie        & 3     & 4,004     & 1,067      & 33.7   \\
        \bottomrule
\end{tabular}}
\caption{Specific configuration comparison of four datasets}
\label{tab:dataset}
\end{table}

\subsubsection{Benchmark and Model}

We leverage the FedPETuning benchmark\cite{zhang2023fedpetuning}, which simulates the application of parameter-efficient fine-tuning in federated learning. Consistent with FedPETuning, we employ the RoBERTa base model~\cite{liu2019roberta} and encapsulate the pre-trained model using the AutoModelForSequenceClassification interface for downstream tasks. The model employs self-supervised learning for pre-training on English data and has been extensively applied in diverse tasks, including sentiment analysis and question-answering. 

\subsubsection{Non-IID Partitioning}

\begin{figure}
  \centering
  \includegraphics[width=0.48\textwidth]{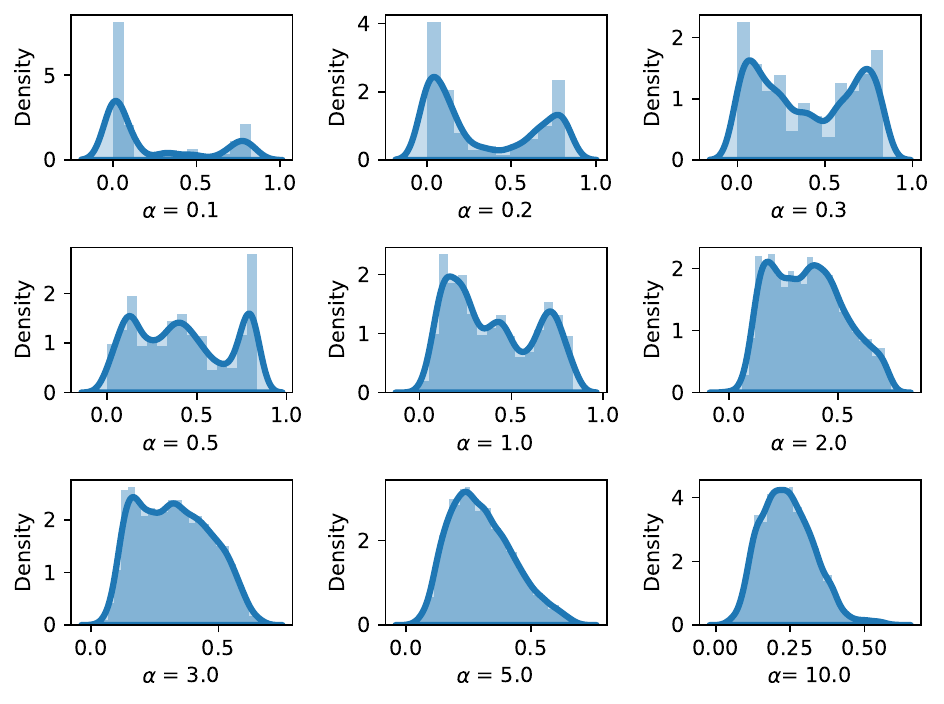}
  \caption{The probability distributions of client data labels under different Dirichlet parameter $\alpha$. Smaller $\alpha$ indicates a higher degree of data heterogeneity.}
  \label{fig:datadist}
\end{figure}

Aligning with the approach outlined in \cite{linfednlp:scheme}, we characterized data heterogeneity by implementing diverse distributions across samples situated within distinct clients. For the Amazon dataset, our selection leaned towards different values of Dirichlet parameter $\alpha$ within a logarithmic scale range (from 0.1 to 10), where smaller $\alpha$ values correspond to non-IID distributions, all chosen based on the Dirichlet distribution. Figure \ref{fig:datadist} precisely illustrates the probability density functions of label distributions among each client pair computed through the Jensen-Shannon distance. Explicitly, as the degree of data heterogeneity increases, the probability density functions exhibit signs of multimodality and sparsity.

\subsection{Data Heterogeneity can Undermine Model Robustness}

\begin{figure}
  \centering
  \includegraphics[width=0.48\textwidth]{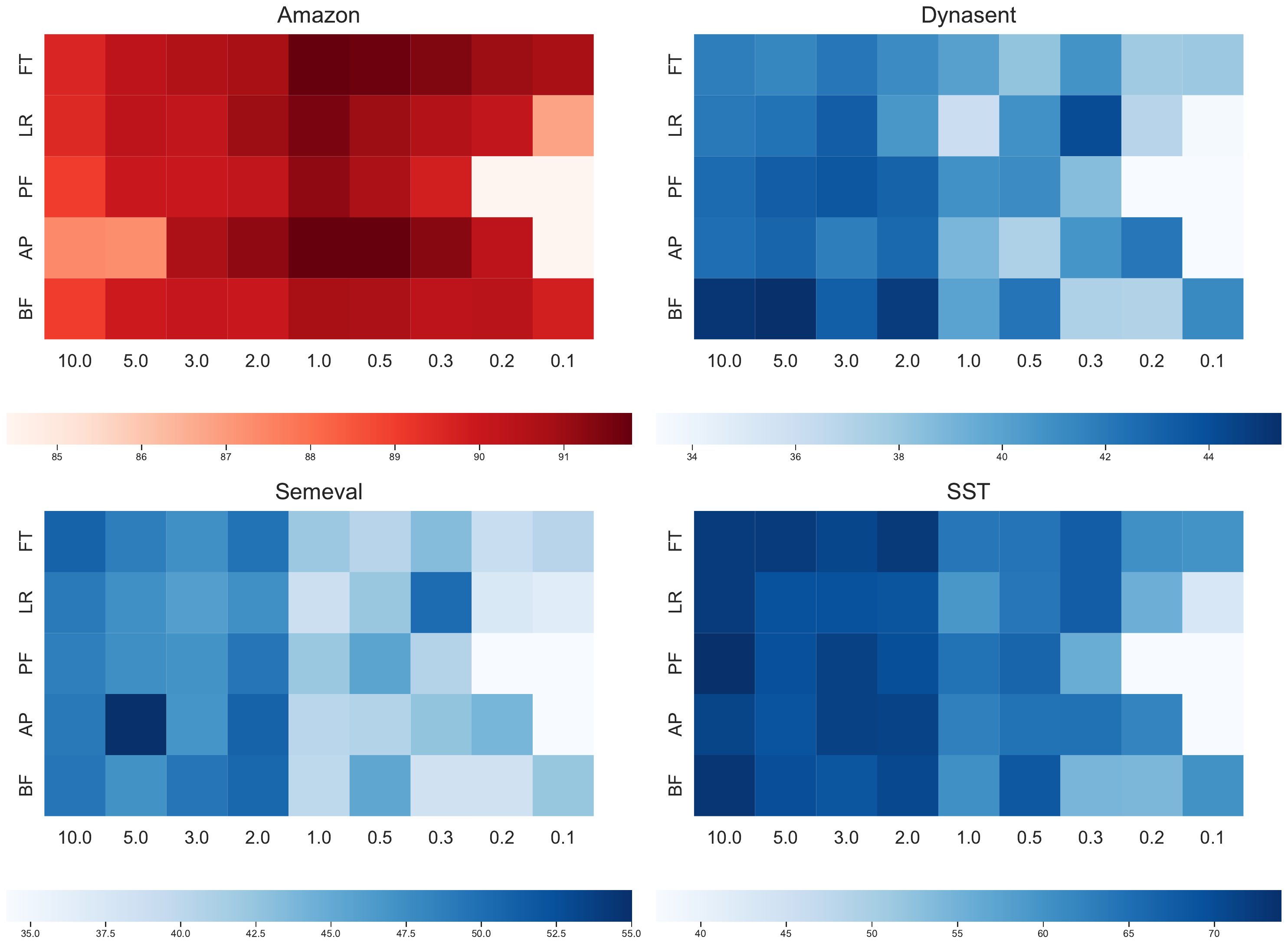}
  \caption{Comparative heatmap of accuracy under different $\alpha$ on Amazon, Dynasent, Semeval, and SST datasets. The vertical axis represents four different fine-tuning methods: Full fine-tuning (FT), LoRA (LR), prefix tuning (PF), adapter tuning (AP), and BitFit (BF). The horizontal axis represents various $\alpha$ values, with increasing data heterogeneity from left to right. The heatmap uses red for datasets with the ID datasets and blue for robust datasets, where darker colors indicate higher accuracy.}
  \label{fig:heat}
\end{figure}

We initially evaluate the OOD robustness of the model by assessing its performance on three robust datasets. As depicted in Figure \ref{fig:heat}, on the ID dataset Amazon, the model's accuracy exhibits a trend of initially increasing and then decreasing with the strengthening of data heterogeneity. Conversely, on the robust datasets Dynasent, SemEval, and SST, the heatmap colors lighten from left to right. This indicates that with increasing data heterogeneity, the model's accuracy significantly decreases on the robust datasets. In terms of specific numerical values, the model's accuracy appears relatively stable on the ID dataset. Taking the full fine-tuning method as an example, as the Dirichlet parameter $\alpha$ varies from 10.0 to 0.1, the accuracy on the ID dataset changes from 89.6\% to 90.8\%. However, on the robust datasets DynaSent, SemEval, and SST, the accuracy decreases from 41.8\% to 37.9\%, 50.9\% to 40.4\%, and 72.3\% to 60.0\%, respectively. The observation of a significant decrease in the model's accuracy on robust datasets reveals that with enhanced data heterogeneity, the model's robustness experiences notable deterioration. This phenomenon also indicates a lack of generalization ability across different distributions for the same task. We further conduct feature space analysis to unearth the underlying reason for the loss in robustness.

\subsubsection{Feature Space Analysis}

\begin{figure}
  \centering
  \includegraphics[width=0.5\textwidth]{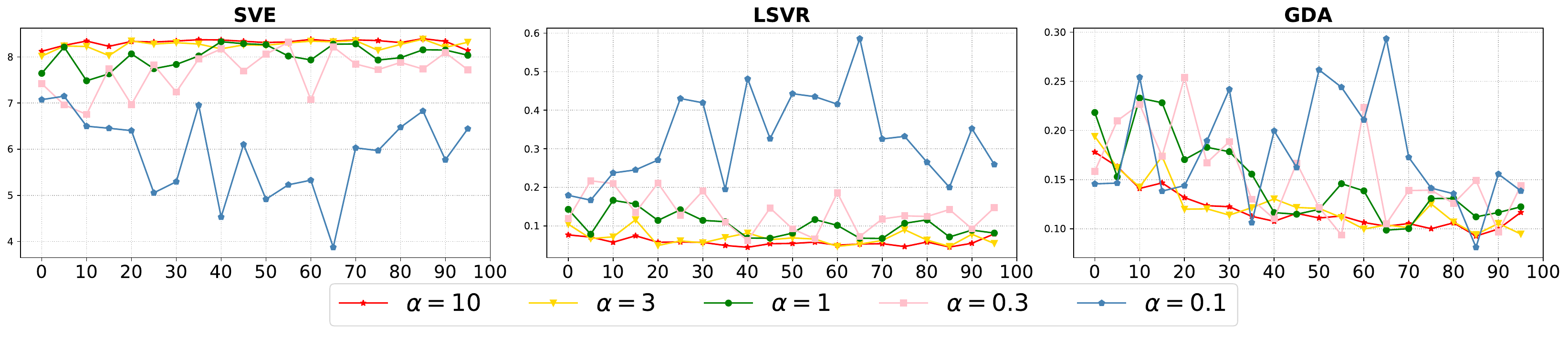}
  \caption{Evolution of three robust indicators with varying data heterogeneity.}
  \label{fig:indicator}
\end{figure}

As shown in Figure \ref{fig:indicator}, we present the changes in three robustness indicators with the model training. Due to space limitations, we chose to showcase the experimental results using the LoRA method, while the variations caused by other fine-tuning methods will be detailed in the Appendix.

With the escalation of data heterogeneity, the SVE value of the model experiences a rapid decrease, which implies that data heterogeneity results in an incomplete capture of data features, thereby impairing the model's representational capacity. Furthermore, as the $\alpha$ value decreases from 10 to 0.1, the LSVR value increases substantially, roughly ascending from 0.06 to around 0.5. This indirectly implies that data heterogeneity diminishes the transferability of models. Additionally, GDA oscillates upward with the augmentation of data heterogeneity, indicating a greater deviation between the pre-trained and fine-tuned models. This suggests that as data becomes more diverse, the adjustments specific to the task might deviate more significantly from the initial pre-trained model, potentially affecting performance. Hence, balancing adapting to task-specific data and retaining the robustness acquired during pre-training becomes pivotal.

\subsection{Diverse PEFT Methods Showcase Varying Degrees of Robustness.}

We employ boxplots to compare the accuracy of different fine-tuning methods visually. The t-test is utilized to explore potential significant differences in accuracy between the full fine-tuning method and other PEFT methods. Smaller values indicate the greater significance of differences.

\begin{figure}
  \centering
  \includegraphics[width=0.48\textwidth]{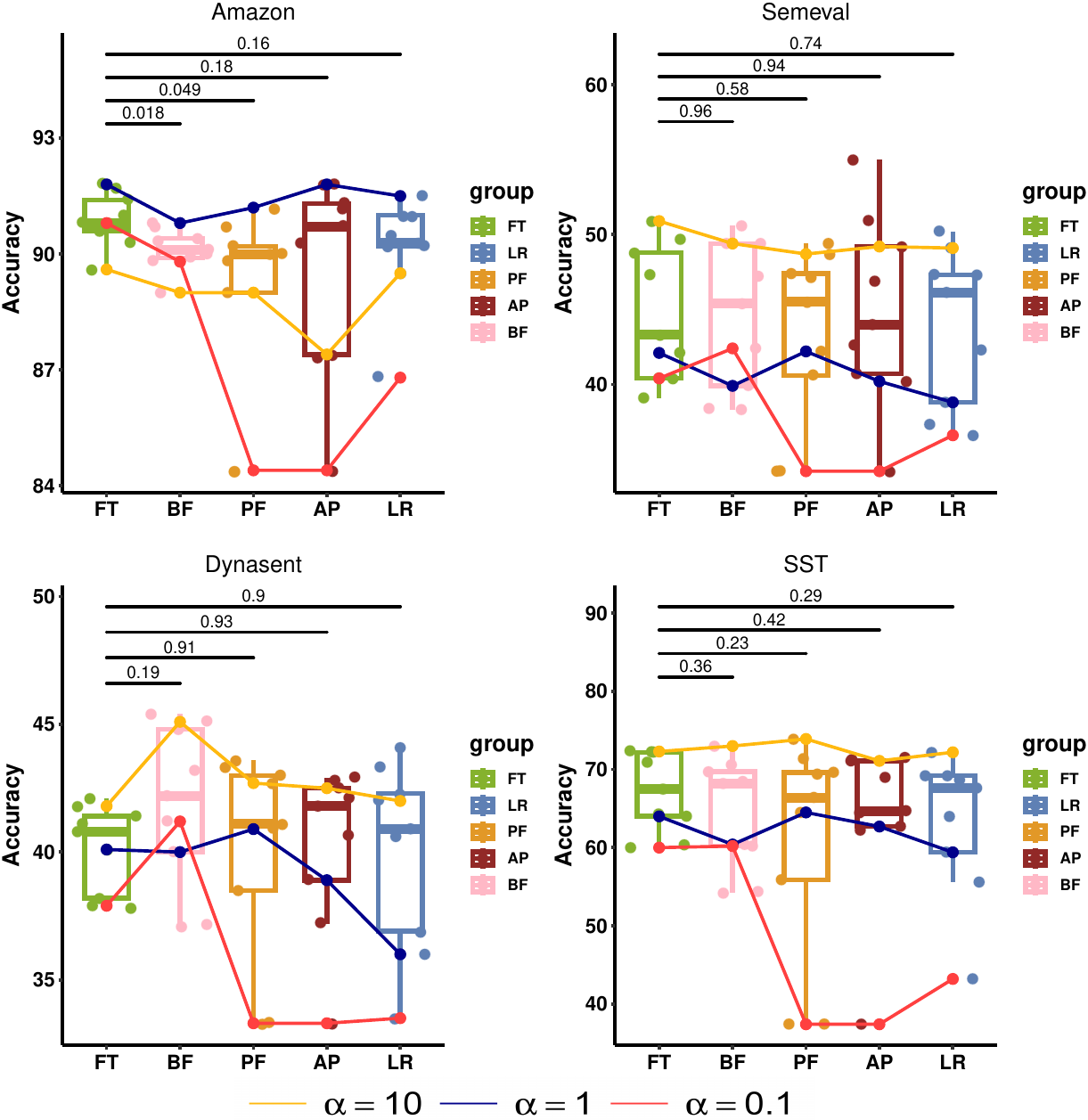}
  \caption{A boxplot comparison illustrates the accuracy difference between the full fine-tuning method and four parameter-efficient fine-tuning methods. Significance differences among the groups are indicated at the top of each subset, with larger values denoting lower differences. The blue, yellow, and red lines represent accuracy at different levels of data heterogeneity.}
  \label{fig:peft}
\end{figure}

As illustrated in Figure \ref{fig:peft}, concerning the ID dataset Amazon, the full fine-tuning method exhibits superior performance compared to other PEFT methods. This advantage arises from the capability to update all layers of the model based on the new task data. However, the advantage of the full fine-tuning method is not as prominent on the robust dataset, where overall accuracy does not exhibit significant variations. Moreover, within the dataset characterized by an $\alpha$ value of 0.1, depicted by the red dashed line, notable disparities in accuracy among distinct PEFT methods on robust datasets become evident. Specifically, BitFit showcases the optimal performance, followed by LoRA, whereas Prefix tuning and Adapter tuning manifest comparatively inferior results. Intriguingly, BitFit exhibits superior performance on the robust dataset compared to the full fine-tuning method. This aligns with recent research findings highlighting BitFit's enhanced generalization performance for tasks involving limited labeled data~\cite{liu2022few}. In our subsequent feature space analysis below, we further delve into the reasons behind this phenomenon.


\subsubsection{Feature Space Analysis}

To comprehensively analyze the variations in robustness within our model, we conducted a comparison of changes in three indicators among different fine-tuning methods using the $\alpha$ value of 0.1. Our observations reveal that BitFit demonstrates a significantly larger SVE and a smaller LSVR compared to the FT method. This finding suggests that the BitFit method effectively captures more data features and possesses greater transferability. Remarkably, our observations indicate an inverse correlation between SVE values and the scale of trainable parameters within the model. Specifically, the utilization of more communication-efficient fine-tuning methods seems to result in higher SVE values. We posit that this phenomenon is primarily attributed to the reduced number of trainable parameters, enabling the classifier layer to capture information within the data comprehensively. This elucidates the significant disparity in accuracy between Bitfit, exhibiting a communication overhead of 12x, and the Prefix method, which demonstrates a substantially higher communication overhead of 190x.

Moreover, in the evaluation of GDA values, the LoRA method demonstrates the highest GDA value, contrasting with the FT method, which exhibits the lowest GDA value. We believe this may be related to the principle of LoRA using a low-rank approximation for fine-tuning. The reduction in matrix rank necessitates a larger GDA in the classifier layer to capture the data features effectively.

\begin{figure}
  \centering
  \includegraphics[width=0.5\textwidth]{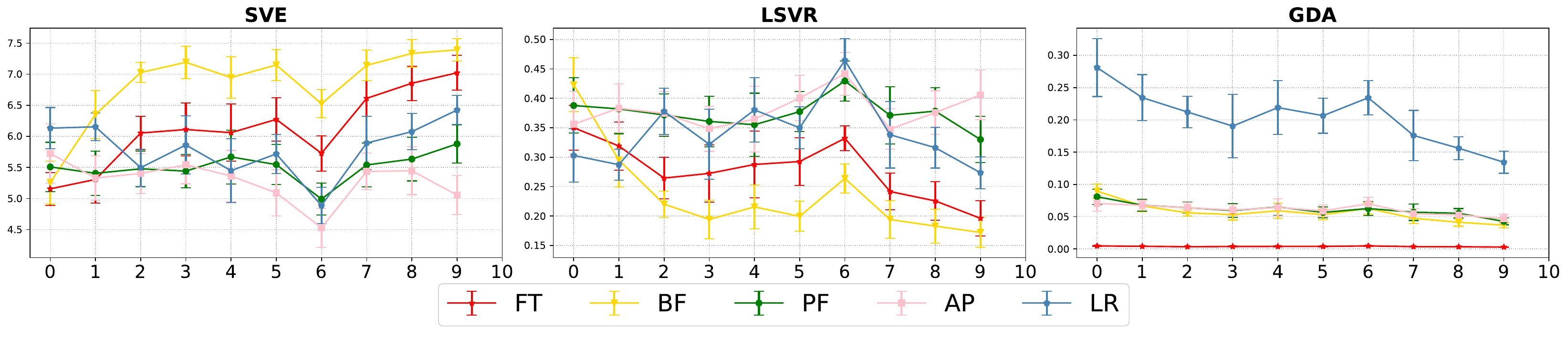}
  \caption{Evolution of three robust indicators with different fine-tuning methods.}
  \label{fig:PEFTindicator}
\end{figure}

\section{The Proposed Method}
\label{section4:robustvector}

\subsection{Problem Formulation}



In federated fine-tuning, a premise is established where different clients $k \in K$ using trainable weights $\theta_{ft}^k$ to participate in joint fine-tuning based on the same task dataset. A central server is responsible for overseeing global model parameters $\tilde{\theta}_{ft}$ instead of raw data. The objective is to ensure that {\em data heterogeneity and the application of PEFT methods do not adversely affect the model's out-of-distribution robustness}.


\subsection{Framework Overview}

In Section 3, an elaborate discussion has been presented on the deterioration of the model's OOD robustness induced by Federated Learning. To tackle this concern, we introduce GNP, a General Noise Projection-based robust fine-tuning algorithm. This algorithm can be integrated as a complementary approach within parameter-efficient fine-tuning methods involving the following three key steps.

\subsubsection{Deriving Robustness Vector from Fine-tuned Models}

Inspired by Chen's research, we project the fine-tuned model $\theta_{ft}$ onto the pre-trained model $\theta_{pre}$ to obtain $P_{\theta_{pre}} \theta_{ft}$. By subtracting the projected model from the fine-tuned model $\theta_{ft}$, we derive the robustness vector $\theta_r$, which characterizes the disparity between the fine-tuned model and the pre-trained model. The computation of $\theta_r$ is illustrated in Equation \ref{eq:robustvector}. Intuitively, this robust vector captures the fundamental changes during the fine-tuning stage. The objective is to leverage this vector to retain the robustness of the pre-trained model Roberta to the greatest extent possible. Simultaneously, the model must enhance its adaptation to the data distribution specific to downstream tasks in the domain.

\begin{equation}
    \theta_{r} = \theta_{ft} - P_{\theta_{pre}} \theta_{ft}
    \label{eq:robustvector}
\end{equation}

\subsubsection{Formulating Value of Fine-tuned Models}

A correlation between the SVE, LSVR, and GDA within the model's feature space and robustness has already been elucidated previously. In this section, we utilize these three indicators to formulate a value model for the fine-tuned model, which represents the weights of $\theta_{ft}$. Intuitively, we expect the model to exhibit enhanced representational capacity, transferability, and minimal deviation from the pre-trained model, corresponding to higher SVE, LSVR, and GDA values, respectively. Subsequently, we formulate the learning weight $\gamma$ for the fine-tuned model in Equation \ref{eq:valuemodel} as follows:

\begin{equation}
     \gamma = \frac{\tau * \mathrm{GDA} * \mathrm{LSVR}}{\mathrm{SVE}},
     \label{eq:valuemodel}
\end{equation}
where $\tau$ is a coefficient for the values of the fine-tuned model. Adjustments to $\tau$ play a crucial role in achieving a balance between model accuracy and robustness.

\subsubsection{Gaussian Noise Injection for Augmenting Robustness}

During the adjustment and update process of the fine-tuned model, the subtraction of robust vectors $\theta_{r}$ is followed by the introduction of Gaussian noise $\theta_{n}$ of the equivalent norm. This inclusion of Gaussian noise serves two critical purposes: Firstly, it contributes to the augmentation of the SVE, facilitating the incorporation of a more comprehensive data structure across broader dimensions. Secondly, the randomness introduced by the noise can generally enhance the model's out-of-distribution robustness. The ultimate computation for updating the fine-tuned model is expressed as follows:

\begin{equation}
    \tilde{\theta}_{ft} = \theta_{ft} - \gamma (\theta_{r}-\theta_{n}).
\end{equation}

\section{Experiment}

We provide empirical analyses concerned with the out-of-distribution robustness of GNP during federated fine-tuning.

\subsection{Implementation Details}

Our experimental setups, encompassing learning rate, scaling factors, and communication rounds, closely mirror those employed in FedPETuning—a benchmark simulating the application of parameter-efficient fine-tuning in federated learning. The FedAvg algorithm is implemented within the FedLab~\cite{zeng2023fedlab}, while the methodology for efficient parameter tuning is drawn from the OpenDelta. In pursuit of universal applicability for our experimental outcomes, we fixed the $\tau$ coefficient constant at 20. All experiments were conducted utilizing 8 NVIDIA GeForce RTX 3090 GPUs. More implementation Details are deferred to the Appendix.

\subsection{Model Accuracy Comparison}

\begin{table*}[]
\centering
\begin{tabular}{cl|rrrrrrrr}
\toprule
\multicolumn{2}{c|}{\multirow{2}{*}{Dataset}} &
  \multicolumn{8}{c}{Method} \\ 
\multicolumn{2}{r|}{} &
  BiFit &
  \multicolumn{1}{c}{BiFit+GNP} &
  Prefix &
  \multicolumn{1}{c}{Prefix+GNP} &
  Adapter &
  \multicolumn{1}{c}{Adapter+GNP} &
  LoRA &
  LoRA+GNP \\ 
  \midrule
\multirow{3}{*}{Amazon} &
  $\alpha$=10.0 &  89.0 &  \multicolumn{1}{c|}{89.1} &  89.0 &  \multicolumn{1}{c|}{89.4} &  87.4 &  \multicolumn{1}{c|}{88.3} &  89.5 &  89.5 \\
 &
  $\alpha$=1.0 &  90.8  & \multicolumn{1}{c|}{91.1} &  91.2 &  \multicolumn{1}{c|}{91.0} & 91.8 & \multicolumn{1}{c|}{91.0} & 91.5 & 91.4 \\
 &
  $\alpha$=0.1 & 89.8 &  \multicolumn{1}{c|}{88.8} &  84.4 & \multicolumn{1}{c|}{84.4} & 84.4 & \multicolumn{1}{c|}{89.4} & 86.8 & 87.5 \\ \hline
\multirow{3}{*}{Dynasent} &
  $\alpha$=10.0 & 45.1 & \multicolumn{1}{c|}{44.9} & 42.7 & \multicolumn{1}{c|}{44.5} & 42.5 & \multicolumn{1}{c|}{43.8} & 42.0 &  43.1 \\
 &
  $\alpha$=1.0 & 40.0 & \multicolumn{1}{c|}{41.1} & 40.9 & \multicolumn{1}{c|}{40.3} & 38.9 & \multicolumn{1}{c|}{41.5} & 36.0 & 40.7 \\
 &
  $\alpha$=0.1 & 41.2 & \multicolumn{1}{c|}{42.4} & 33.3 & \multicolumn{1}{c|}{33.3} &  33.3 & \multicolumn{1}{c|}{40.9} & 33.5 & 43.6 \\ \hline
\multirow{3}{*}{SemEval} &
  $\alpha$=10.0 & 49.4 & \multicolumn{1}{c|}{49.6} & 48.7 & \multicolumn{1}{c|}{49.5} & 49.2 & \multicolumn{1}{c|}{50.5} & 49.1 & 53.0 \\
 &
  $\alpha$=1.0 & 39.9 & \multicolumn{1}{c|}{43.1} & 42.2 & \multicolumn{1}{c|}{43.4} & 40.2 & \multicolumn{1}{c|}{47.9} & 38.8 & 46.0 \\
 &
  $\alpha$=0.1 & 42.4 & \multicolumn{1}{c|}{43.6} & 34.2 & \multicolumn{1}{c|}{34.2} & 34.2 & \multicolumn{1}{c|}{46.5} & 36.6 & 41.2 \\ \hline
\multirow{3}{*}{SST} &
  $\alpha$=10.0 & 73.0 & \multicolumn{1}{c|}{72.4} & 73.9 & \multicolumn{1}{c|}{72.9} & 71.1 & \multicolumn{1}{c|}{73.1} & 72.2 & 71.8 \\
 &
  $\alpha$=1.0 & 60.4 & \multicolumn{1}{c|}{64.6} & 64.5 & \multicolumn{1}{c|}{66.4} & 62.7 & \multicolumn{1}{c|}{69.8} & 59.4 & 67.9 \\
 &
  $\alpha$=0.1 &
  60.2 & \multicolumn{1}{c|}{60.2} & 37.4 & \multicolumn{1}{c|}{37.4} & 37.4 & \multicolumn{1}{c|}{63.9} & 43.2 & 54.8 \\ 
  \bottomrule
\end{tabular}
\caption{Algorithm Comparison: Testing Accuracy.}
\label{tab:accuracy}
\end{table*}

We integrated our General Noisy Projection-based robust fine-tuning algorithm (GNP) within four distinct methods, namely BitFit (BF), Prefix Tuning (PF), Adapter Tuning (AP), and LoRA (LR). Performance results for GNP, evaluated across diverse data distributions within these four datasets, are presented in Table \ref{tab:accuracy}.


In the ID dataset on Amazon, diverse fine-tuning methods for GNP-based algorithms exhibit marginal deviation from the initial methodology. Notably, characterized by $\alpha$ of 0.1, the GNP-based algorithm attains a 5\% enhancement in the accuracy of the Adapter tuning. In light of an intensified emphasis on OOD robustness, the performance accuracy of the GNP-based methodology on the Amazon dataset presents persuasive evidence affirming its ability to sustain the model's efficacy on the ID dataset, all while avoiding any compromise to its overarching performance.

In the case of the Dynasent, SemEval, and SST robust datasets, the GNP method exhibits more stable and superior results as $\alpha$ decreases from 10.0 to 0.1. The alignment with our expectations lies in effectively controlling data heterogeneity, aiming to mitigate any adverse effects on model robustness. For instance, when $\alpha$ is set to 0.1, GNP on the Dynasent dataset shows an improvement from 33.5\% to 43.6\% compared to the LoRA method. On the SemEval dataset, the comparison with the Adapter tuning method increases from 34.2\% to 46.5\%. Similarly, the comparison with the Adapter tuning method on SST rises from 37.4\%  to 63.9\%.

While GNP-based algorithms typically outperform the original approach, our method demonstrates diverse enhancement effects when applied to various PEFT methods. Notably, the improvements achieved through our method in Adapter tuning and LoRA significantly surpass those observed in Prefix tuning. The reasons behind this phenomenon are elucidated in the following discussion.


\subsection{Comparison of Three Robustness Indicators}

\begin{figure}
  \centering
  \includegraphics[width=0.4\textwidth]{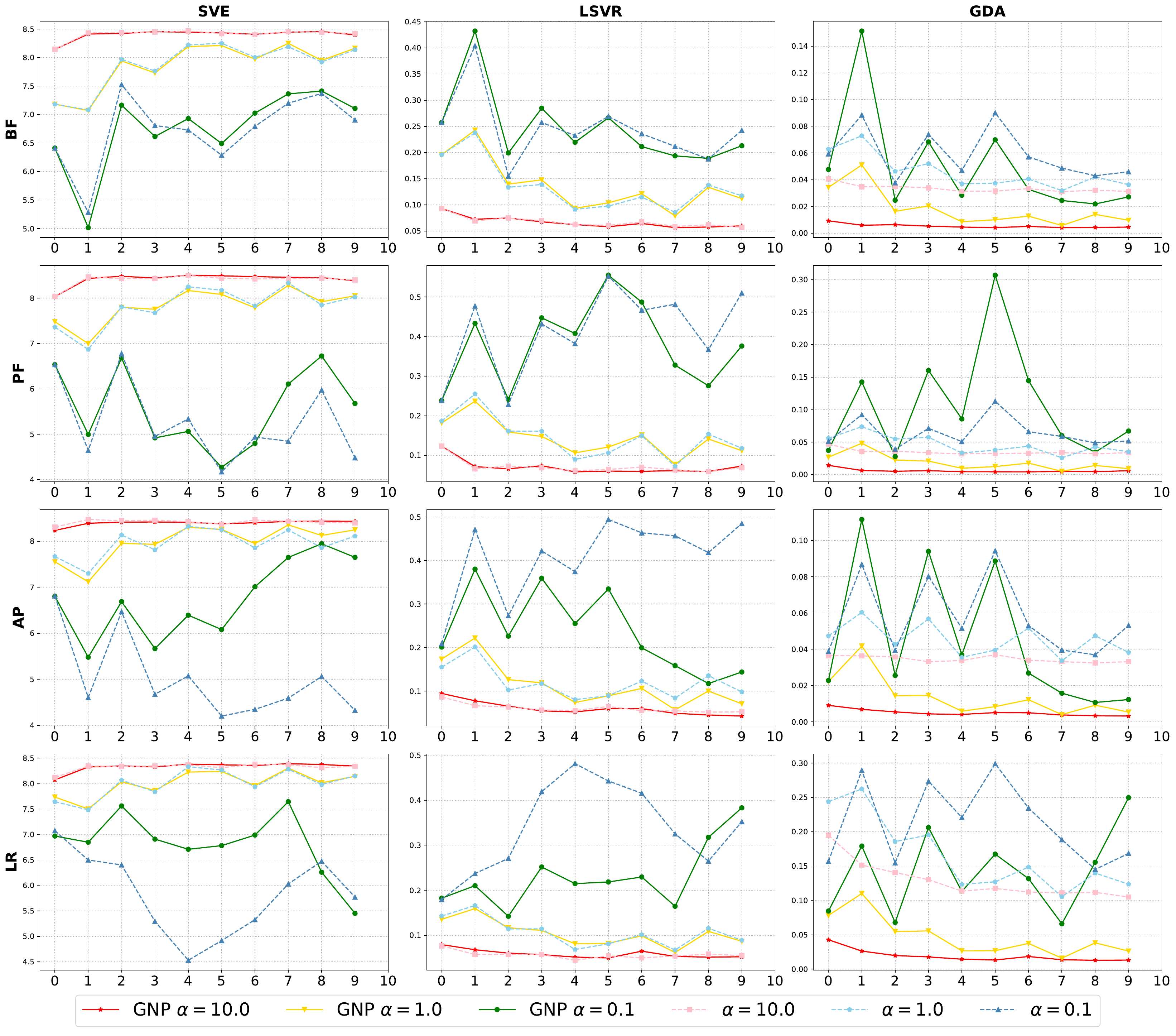}
  \caption{Comparison of three robustness indicators under the GNP method across various parameter-efficient fine-tuning methods in three distinct data heterogeneity scenarios, quantified with $\alpha$ of 10.0, 1.0, and 0.1. Solid lines depict the application of the GNP method.}
  \label{fig:robustindcator}
\end{figure}

As illustrated in Figure \ref{fig:robustindcator}, the robustness indicators of the GNP-based methods remain within manageable bounds when considering $\alpha$ values of both 10 and 0.1. Notably, the application of GNP is more effective when $\alpha$ is set to 0.1. Given the absence of prior information regarding data heterogeneity, we assert that our method exhibits greater generality.

When faced with a higher degree of data heterogeneity, the model experiences a substantial decline in SVE values. In this context, the application of GNP proves effective in enhancing the model's SVE values, demonstrating notable superiority, especially in Adapter tuning and LoRA. Furthermore, with an increase in the number of model training iterations, GNP-based methods successfully diminish the LSVR values, showcasing improved transferability. Notably, in the Adapter tuning method, as the $\alpha$ value decreases from 1.0 to 0.1, the LSVR values gradually approach those observed at $\alpha$ 1.0. This observation suggests that our method indeed exhibits superior adaptability to disparate data distributions, particularly evident when $\alpha$ is set to 0.1.

In terms of the fluctuation in GDA values, significant reductions are observed in BitFit, Adapter tuning, and LoRA, as depicted by the solid line. This implies that the GNP-based methods effectively introduce mechanisms capable of reducing the model's deviation. However, Prefix tuning exhibits a larger fluctuation, correlating with the accuracy performance observed in Table \ref{tab:accuracy}. We attribute this phenomenon to the decline of SVE values reaching a critical threshold during model training in the context of Prefix tuning methods. Since our approach to enhancing SVE values involves introducing general Gaussian noise, although this facilitates the model to traverse additional dimensions, the efficacy is finite and fails to produce satisfactory results when SVE descends to the threshold. Moreover, the excessive addition of Gaussian noise can also impact the model's accuracy. Based on the observations above, an explanation is provided for why the introduction of GNP leads to diverse OOD robustness performance across different fine-tuning methods.

\section{Conclusion}


In this study, we initially disclose that federated learning has the risk of distorting model features and compromising out-of-distribution robustness when fine-tuning. To alleviate the negative impact of data heterogeneity and communication limitations, we propose a General Noise Projection-based robust algorithm, denoted as GNP. Specifically, by incorporating three robustness indicators, robust vectors are meticulously crafted from the fine-tuned model to inherit robustness from the pre-trained model. Subsequently, Gaussian noise is introduced to enhance the model's representative capacity, thus improving robustness without compromising model performance. Compared with other parameter-efficient fine-tuning methods, our approach maintains commendable robustness without sacrificing accuracy in the ID distribution.

\appendix

\section*{Appendix}

\section{Experimental Setting}

This section provides a detailed overview of the experimental setup, encompassing the dataset, model architecture, federated learning settings, and code details.

\subsection{Dataset}

In alignment with the benchmark FedPETuning, we designate the validation dataset within the Amazon dataset as the test dataset and partition a dataset segment for validation purposes. Furthermore, we select three robust test datasets derived from DynaSent, SemEval, and SST. The detailed descriptions of the four datasets are provided as follows:

\begin{itemize}
    \item \textbf{Amazon~\cite{amazon}:} The dataset encompasses product reviews from Amazon up to 2014, including reviews, product metadata, and linkages. It provides a comprehensive repository of information about consumer evaluations and product details.
    \item \textbf{DynaSent~\cite{potts2020dynasent}:} The Dynamic Sentiment Analysis Dataset is a ternary task for dynamic sentiment analysis. It primarily focuses on explicit, context-independent expressions of emotion. This dataset can be used for emotion recognition, sentiment classification, and sentiment analysis.
    \item \textbf{SemEval~\cite{semeval}:} This dataset is meticulously curated for sentiment analysis tasks conducted on Twitter. It encompasses five distinct subtasks designed to facilitate in-depth exploration and comprehension of emotion analysis from multiple perspectives.
    \item \textbf{SST~\cite{sst}:} The Stanford Sentiment Treebank is a pioneering corpus that includes fully labeled parse trees, allowing for a comprehensive analysis of sentiment effects. This dataset consists of sentence-level movie reviews sourced from the Rotten Tomatoes website.
\end{itemize}

\subsection{Model Architectures}

The Roberta-base model employs the masked language modeling (MLM) objective and undergoes pre-training on 160GB of English text in a self-supervised fashion. The training involves 1024 V100 GPUs for 500,000 steps, using a batch size of 8,000 and a sequence length of 512. The optimizer utilized is Adam with $\beta_1=0.9$ and $\beta_2=0.98$. Other vital parameters include a weight decay of 0.01, a learning rate warm-up for 24,000 steps, and a linear decay of the learning rate after that. The intricate particulars of the model have been made publicly available on the Hugging Face\footnote{\url{https://huggingface.co/roberta-base}}.

\subsubsection{Trainable Weights}


The essence of the parameter-efficient fine-tuning method lies in freezing most of the model while selecting a small portion for training updates to achieve communication efficiency. Table \ref{tab:booktabs} illustrates the percentage of trainable parameters for different fine-tuning methods, where a smaller percentage indicates higher communication efficiency.

\begin{table}
    \centering
    \begin{tabular}{lrr}
        \toprule
        Algorithms  & Trainable Parameters & Ration \\
        \midrule
        Full Fine-tuning      & 124647939          & 100.00\%         \\
        BitFit                &   657411           & 0.53\%        \\
        Prefix Tuning         &   1031424          & 8.27\%        \\
        Adapter Tuning        &   1496643          & 1.20\%       \\
        LoRA                  &   887811           & 0.71\%       \\
        \bottomrule
    \end{tabular}
    \caption{Table of Notations}
    \label{tab:booktabs}
\end{table}



\subsection{Federated Learning Settings}

In a federated fine-tuning setting, a supposition is made wherein a central server is tasked with the management of global model parameters $\tilde{\theta}_{ft}$, along with the accumulation of trainable weights $\theta^k_{ft}$ from $\mathcal{K}$ clients. Each client possesses a distinct privacy dataset $\mathcal{D}_k$ within their respective domain, wherein the variations in $\mathcal{D}_k$ are contingent upon distinct data distributions governed by the Dirichlet parameter $\alpha$.
\begin{equation}
    \mathrm{min}\mathcal{L}(\theta^1_{ft},\dots,\theta^K_{ft};p)=\Sigma_{k=1}^Kp_k\mathbb{E}_{(x,y)\sim D_k}[\mathcal{L}_k(\theta^k_{ft};(x,y)],
\end{equation}

\subsection{Code Details}


For the reviewer's convenience, we have utilized an anonymous GitHub repository to provide comprehensive code details. Our repository will be available at {\em GNP}\footnote{\url{https://anonymous.4open.science/r/GNP-00F4/README.md}}.

\subsection{Notations}

We introduce the notation for the parameters utilized in this paper in Table \ref{tab:notation}.

\begin{table}
    \centering
    \begin{tabular}{ll}
        \toprule
        Notation  & Trainable Parameters  \\
        \midrule
        SVE                              & Singular Value Entropy         \\
        LSVR                             &   Largest Singular Value Ratio           \\
        GDA                              &   Gradient Deviation Angle         \\
        $\theta_{\mathcal{C}}$           &   The weight layer of the classifier in fine-tuned model          \\
        $\theta_{ft}$                    &   The trainable weights in fine-tuned model          \\
        $\theta_{r}$                     &   Robust vectors calculated by Eq.(\ref{eq:robustvectorall})          \\
        $\theta_{n}$                     &   Gaussian noise with the same shape as $\theta_{r}$          \\
        $K$                              &   Number of clients          \\
        $\cup_{k=1}^K D_k$               &   The dataset constructed by $K$ clients          \\
        $\eta$                           &   Learning rate          \\
        $\gamma$                         &   Robust evaluation metric calculated by Eq.(\ref{eq:valuemodel})        \\
        $E$                              &   Local steps         \\
        $c$                              &   Sampling rate          \\
        $T$                              &   Communication rounds         \\
        \bottomrule
    \end{tabular}
    \caption{Comparison of Trainable Parameters Across Different Fine-Tuning Algorithms}
    \label{tab:notation}
\end{table}

\section{Algorithms}

In this section, we offer a comprehensive exploration of algorithmic details, including the specific implementation and pseudocode.

\subsection{Matrix Projection}


In machine learning, projection is a valuable tool to enhance our understanding of spatial transformations. Equation \ref{eq:projection} elaborates on the computational process involved in projecting matrix $\mathbf{B}$ onto matrix $\mathbf{A}$. More precisely, it entails the initial computation of the projection matrix, followed by mapping the original matrix to a subspace.

\begin{equation}
    \label{eq:projection}
    P_{\mathbf{A}}(\mathbf{B}) = \frac{\mathbf{A}\mathbf{A}^T}{\mathbf{A}^T \mathbf{A}} \mathbf{B}^T
\end{equation}

\subsection{Robust Vectors}


By leveraging the properties of projection and drawing inspiration from the concept of Robust Weight Signatures\cite{cai-et-al:scheme}, we project matrix $\theta^{t+1}_{ft}$ onto matrix $\theta^{t}_{ft}$, designating the obtained result as standard dataset-fitting knowledge. Intuitively, $\theta^{t+1}_{ft}$ encapsulates the complete knowledge acquired by clients in the dataset at $t$, encompassing robust knowledge. Following this, we decouple {\em robust vectors} by subtracting the projection result from $\theta^{t+1}_{ft}$, as outlined in Equation \ref{eq:robustvectorall}.

\begin{equation}
    \theta_{r}^{t+1} = \theta^{t+1}_{ft}-P_{\theta^{t}_{ft}}(\theta^{t+1}_{ft})
    \label{eq:robustvectorall}
\end{equation}

Due to our scenario being situated in domains with scarce and sensitive datasets, such as network security and healthcare, we do not assume the existence of a pre-trained model trained exclusively on a clean dataset, in contrast to \cite{cai-et-al:scheme}. This paper exploits information from the previous time step to complement the pre-trained model for simplicity. Importantly, this approach to augmenting pre-trained model knowledge is highly scalable, accomplished by modifying the base projection space.

\subsection{Robust Indicators}


Section \ref{section3.1 robust indicator} provides a concise overview of the intrinsic representation of robust indicators within the model's feature space. This section subsequently elaborates on the comprehensive methodology employed to compute robust indicators computationally efficiently.

We utilize the weight layer of the model's classifier to depict the global model for capturing robust indicators. This decision is motivated by two primary reasons: firstly, confining computation to the classifier's weight layer significantly reduces computational costs, supported by research indicating the adequacy of shallow model representations; secondly, the classifier layer, operating as a generic linear layer, enables the consistent computation of robust indicators across the diverse parameter-efficient fine-tuning models.

At time $t$, the weight layer of the model's classifier is denoted as $\theta_{t}^{\mathcal{C}} \in \mathcal{R}^{m \times n}$. The weight space undergoes Singular Value Decomposition (SVD) as described in the equation \ref{eq: classifier SVD}.



\begin{equation}
    \label{eq: classifier SVD}
    \theta^{t}_{\mathcal{C}}=\mathbf{U_{t}\Sigma_{t}}\mathbf{V}_{t}^{\top}.
\end{equation}

Here, the $\Sigma_{t} \in \mathcal{R}^{m \times n}$ represents a diagonal singular value matrix $\{\sigma_1^t,...,\sigma_D^t\}$, where $\sigma_1^t$ represents the maximum eigenvalue at $t$. 



The calculation of Singular Value Entropy (SVE) and Largest Singular Value Ratio (LSVR) is explicated in equations \ref{eq: classifier SVE} and \ref{eq: classifier LSVR}, respectively. Specifically, the time complexity of this algorithm is predominantly governed by the SVD. Typically, for $\theta_{t}^{\mathcal{C}} \in \mathcal{R}^{m \times n}$, the time complexity of its SVD is approximately $\mathcal{O}(\min(m^2n, mn^2))$. The algorithm's complexity is contingent upon the scale of the input matrix. By exclusively leveraging the weight layer of the model's classifier, we can markedly augment the algorithm's efficiency and versatility.

\begin{equation}
    \label{eq: classifier SVE}
    \mathrm{SVE}^t=-\sum_{i=1}^D\frac{\sigma_i^t}{\sum_{j=1}^D\sigma_j^t}\log\frac{\sigma_i^t}{\sum_{j=1}^D\sigma_j^t}
\end{equation}

\begin{equation}
    \label{eq: classifier LSVR}
    \mathrm{LSVR}^t=\frac{\sigma_{1}^t}{\sum_{i=1}^{D}\sigma_{i}^t}\
\end{equation}

\begin{algorithm}[tb]
    \caption{Training procedure of GNP}
    \label{alg:algorithm}
    \textbf{Input}: $\theta_0$: parameters of the pre-trained model, $T$: communication rounds, $K$: number of clients, $\cup_{k=1}^K D_k$: dataset, $\eta$: learning rate, $c$: sampling rate, $E$: local steps\\
    \textbf{Output}: Your algorithm's output
    \begin{algorithmic}[1]
        \STATE$\textbf{Server executes:}$
        \STATE $\tilde{\theta^1} \leftarrow \theta^0$ \\
        \STATE $\gamma \leftarrow 1.0 $\\
        \FOR{$t = 1,...,T$}
            \STATE Select $m = c \times K$ clients \\
            \textbf{for} each client $k = 1,...,m$ in parallel \textbf{do}
                \STATE \quad Send $\tilde{\theta}^t_{ft}$ to client $k$
                \STATE \quad $\theta^{t+1}_k \leftarrow \text{ClientUpdate}(k, \tilde{\theta^t})$
            \IF {t \% 10 == 0}
            \STATE Calculate robust indicators by Eq.(\ref{eq: classifier SVE}, \ref{eq: classifier LSVR}, \ref{eq: classifier GDA})
            \STATE Update $\gamma$ by Eq.(\ref{eq:valuemodel})
            \ENDIF   
            \STATE $\theta^{t+1}_{ft} = \sum_{k=1}^{m} \frac{|D_{{\mathcal{B}}}^k|}{\sum_{k=1}^{m}|D_{{\mathcal{B}}}^k|} \theta^{t+1}_k$
            \STATE Calculate robust vectors $\theta_r^{t+1}$ by Eq.(\ref{eq:robustvectorall})
            \STATE Generate Gaussian noise $\theta_n^{t+1}$ 
            \STATE $\tilde{\theta}^{t+1}_{ft} = \theta^{t+1}_{ft} - \gamma (\theta_r^{t+1}-\theta_n^{t+1})$
        \ENDFOR
        \STATE$\textbf{ClientUpdate($k, \tilde{\theta^t_{ft}}$):}$
            \FOR{each local epoch $e = 1,...,E$}
                \STATE $D^{{\mathcal{B}}}_k \leftarrow$ sample a minibatch $\mathcal{B}$ from $D_k$
                \STATE $\theta^{t+1}_k  \leftarrow  \tilde{\theta}^t_{ft} - \eta \nabla(\tilde{\theta}^t_{ft},D^{{\mathcal{B}}}_k)$
            \ENDFOR
        \STATE Return $\theta^{t+1}_k$ to server
    \end{algorithmic}
\end{algorithm}


The computation of Gradient Deviation Angle (GDA) is primarily defined in equation \ref{eq: classifier GDA}: 
\begin{equation}
    \label{eq: classifier GDA}
    \mathrm{GDA}^t = \frac{\theta^{t}_{\mathcal{C}}-P_{\theta^{t-1}_{\mathcal{C}}}(\theta^{t}_{\mathcal{C}})}{\theta^{t}_{\mathcal{C}}},
\end{equation}
where the operation of $P$ is explicitly specified in equation \ref{eq:projection}. The principal complexity of the entire algorithm arises from the matrix multiplication operation, exhibiting a time complexity of $\mathcal{O}(\min(m^2n, mn^2))$, which aligns with the complexities associated with SVE and LSVR.

\subsection{Algorithm Pseudocode}


Algorithm 1 outlines the pseudocode for the training procedure of the GNP algorithm. The server execution is encapsulated in Steps 1-16. Specifically, in each communication round $t$, the server initiates by randomly selecting $m$ clients to send fine-tuned models (Steps 5-7). Additionally, a weighted aggregation is performed based on the proportion of client datasets to obtain model $\theta^{t+1}_{ft}$ (Step 12). Subsequently, robust vectors and Gaussian noise are computed for updating the fine-tuned model $\tilde{\theta}^{t+1}_{ft}$ (Steps 13-14). Moreover, robust indicators are computed every 10 rounds to update $\gamma$ (Steps 8-10). The specifics of client execution closely resemble the federated learning setting, as depicted in Steps 17-21.


\section{Additional Plots}

Due to space constraints, this section serves as a supplement, offering additional plots.

\subsection{Feature Space Analysis}
\begin{figure}
  \centering
  \includegraphics[width=0.5\textwidth]{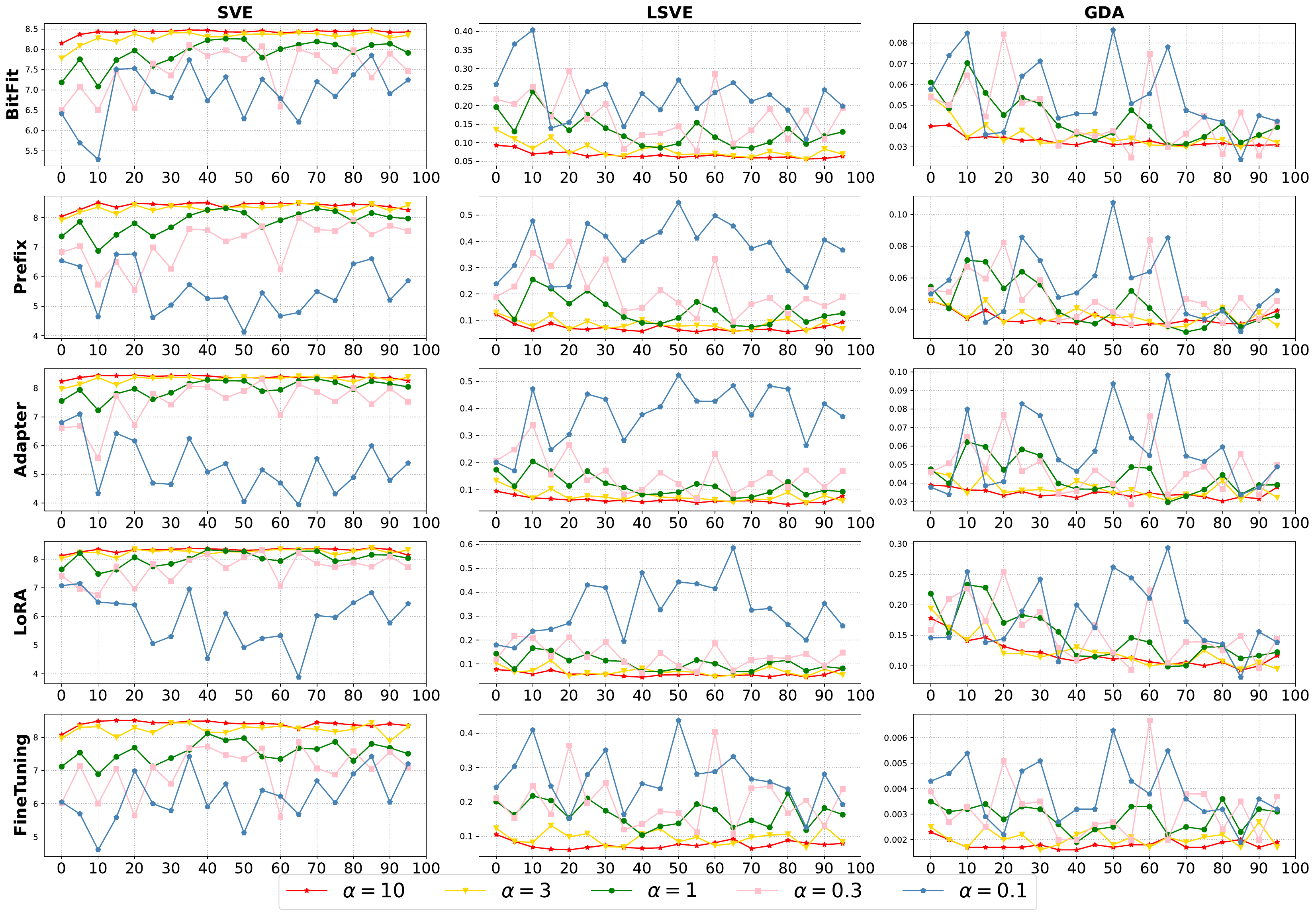}
  \caption{Evolution of three robust indicators with varying data heterogeneity. Horizontal Axis: Communication Rounds. Vertical Axis: Values of Robust Indicators.}
  \label{fig:robustindcatorall}
\end{figure}

Figure \ref{fig:robustindcatorall} serves as a complement to the previously presented Figure \ref{fig:indicator}, illustrating the varied trends of values of three robust indicators for different parameter-efficient fine-tuning methods, namely BitFit (BF), prefix tuning (PF), adapter tuning (AP), LoRA (LR), and Full fine-tuning (FT), over varying communication rounds. Upon visual inspection, it is evident that despite variations in fine-tuning methods, the three robust indicators exhibit a consistent trend in response to increased data heterogeneity.

\subsection{Client Numbers}

\begin{figure}
  \centering
  \includegraphics[width=0.5\textwidth]{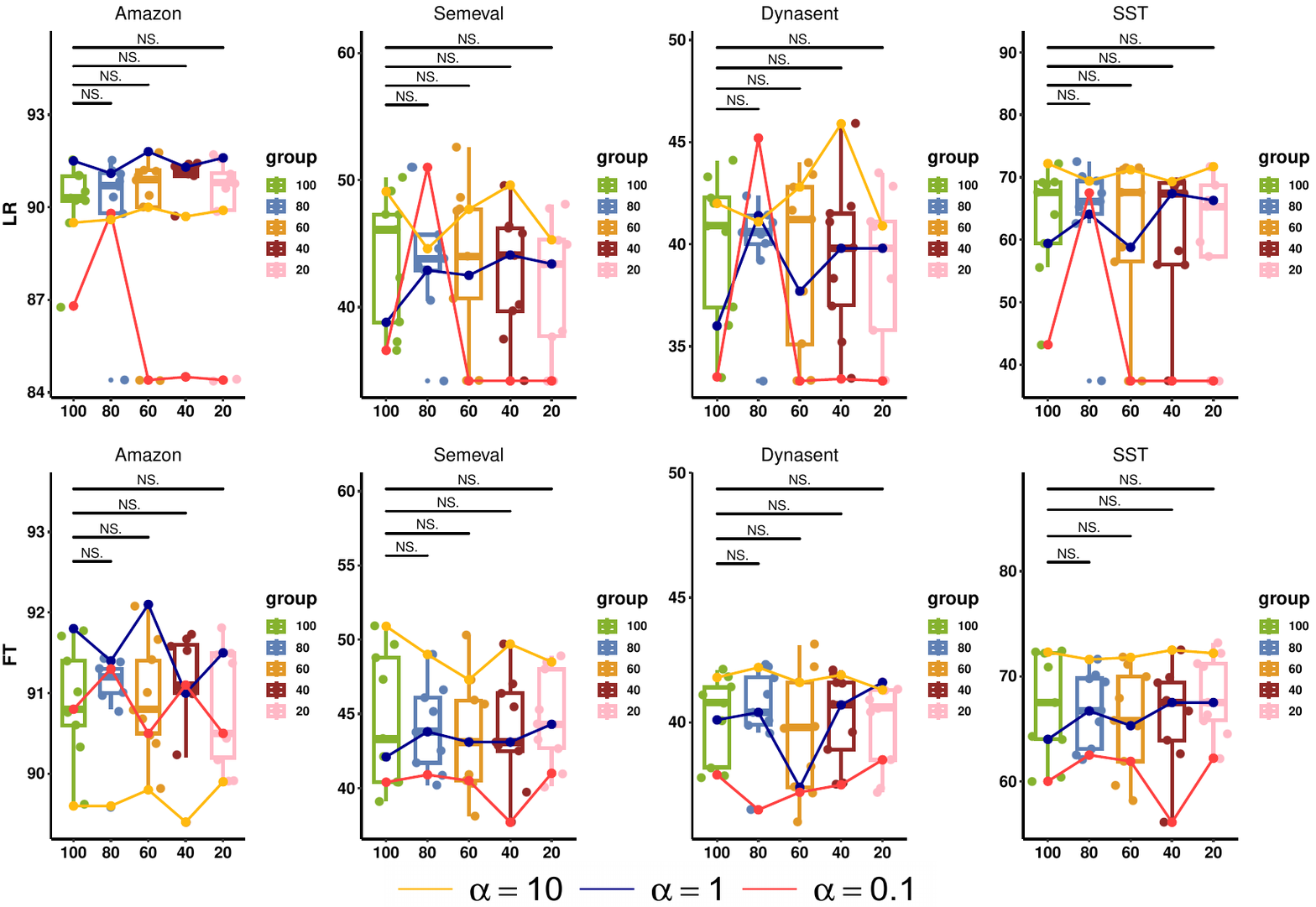}
  \caption{A boxplot comparison of the accuracy between different numbers of clients. The first column displays results for LoRA, while the second column presents results for Full fine-tuning. Marked at the top of each subset indicates the significant differences among the groups, with NS denoting non-significance($p > 0.05$). Horizontal Axis: Numbers of Clients. Vertical Axis: Accuracy.}
  \label{fig:clientsnum}
\end{figure}

We employed the t-test in boxplots to investigate potential significant differences in accuracy among different client numbers, as illustrated in Figure \ref{fig:clientsnum}. We evenly partitioned the dataset in the experiment based on the number of clients. We included 10 clients for model aggregation in each communication round to ensure fairness. It is discernible that, under constant dataset size, varying numbers of clients do not exhibit significant impact accuracy.

\subsection{Accuracy}


In this paper, our objective is to enhance out-of-distribution robustness without compromising the accuracy of the in-distribution (ID) dataset. Consequently, we aim for minimal impact on the ID dataset's accuracy after applying the GNP algorithm. Figure 9 illustrates the accuracy comparison when applying the GNP algorithm on the ID dataset Amazon. As depicted in the figure, under conditions where $\alpha$ is 10.0 and 1.0, applying GNP does not adversely affect the model's accuracy. Conversely, in the presence of solid data heterogeneity ($\alpha$ is 0.1), the GNP algorithm positively impacts accuracy in both Adapter tuning and LoRA. The experimental results indicate that our approach is practical and versatile, as it does not compromise accuracy on the Amazon dataset.

\begin{figure}
  \centering
  \includegraphics[width=0.45\textwidth]{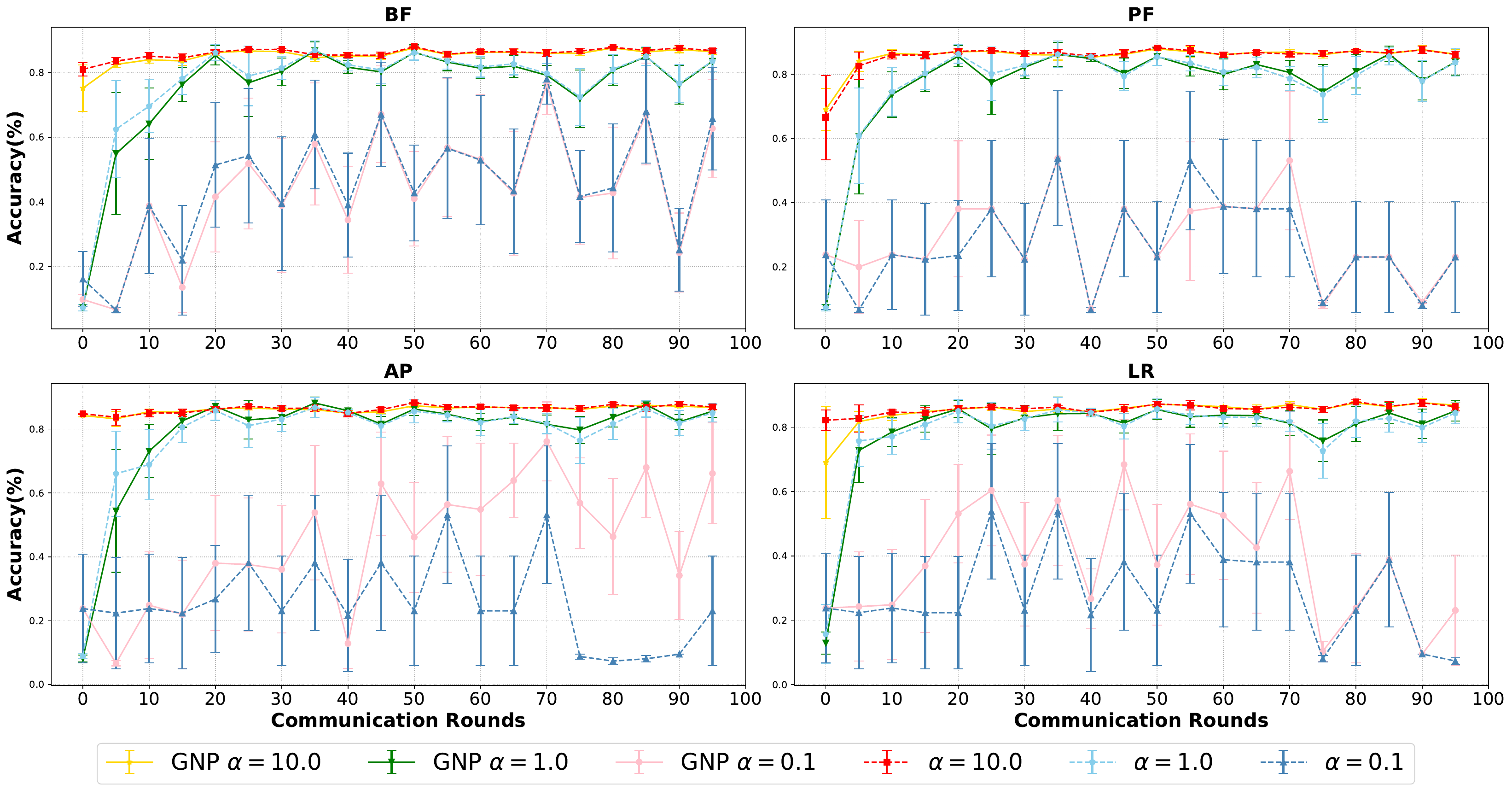}
  \caption{Comparative accuracy of applying the GNP on the ID dataset Amazon. The solid line represents the application of GNP, while the dashed line represents its absence. Horizontal Axis: Communication Rounds. Vertical Axis: Accuracy.}
  \label{fig:accuracy}
\end{figure}




\nocite{*}
\bibliographystyle{named} 
\bibliography{ijcai23} 

\begin{thebibliography}{}

\bibitem[\protect\citeauthoryear{Abelson \bgroup \em et al.\egroup }{1985}]{abelson-et-al:scheme}
Harold Abelson, Gerald~Jay Sussman, and Julie Sussman.
\newblock {\em Structure and Interpretation of Computer Programs}.
\newblock MIT Press, Cambridge, Massachusetts, 1985.

\bibitem[\protect\citeauthoryear{Baumgartner \bgroup \em et al.\egroup }{2001}]{bgf:Lixto}
Robert Baumgartner, Georg Gottlob, and Sergio Flesca.
\newblock Visual information extraction with {Lixto}.
\newblock In {\em Proceedings of the 27th International Conference on Very Large Databases}, pages 119--128, Rome, Italy, September 2001. Morgan Kaufmann.

\bibitem[\protect\citeauthoryear{Brachman and Schmolze}{1985}]{brachman-schmolze:kl-one}
Ronald~J. Brachman and James~G. Schmolze.
\newblock An overview of the {KL-ONE} knowledge representation system.
\newblock {\em Cognitive Science}, 9(2):171--216, April--June 1985.

\bibitem[\protect\citeauthoryear{Brown \bgroup \em et al.\egroup }{2020}]{brown-et-al:scheme}
Tom Brown, Benjamin Mann, Nick Ryder, Melanie Subbiah, Jared~D Kaplan, Prafulla Dhariwal, Arvind Neelakantan, Pranav Shyam, Girish Sastry, Amanda Askell, et~al.
\newblock Language models are few-shot learners.
\newblock {\em Advances in neural information processing systems}, 33:1877--1901, 2020.

\bibitem[\protect\citeauthoryear{Cai \bgroup \em et al.\egroup }{2023}]{cai-et-al:scheme}
Ruisi Cai, Zhenyu Zhang, and Zhangyang Wang.
\newblock Robust weight signatures: Gaining robustness as easy as patching weights?
\newblock {\em arXiv preprint arXiv:2302.12480}, 2023.

\bibitem[\protect\citeauthoryear{Chen \bgroup \em et al.\egroup }{2019}]{chen2019transferability}
Xinyang Chen, Sinan Wang, Mingsheng Long, and Jianmin Wang.
\newblock Transferability vs. discriminability: Batch spectral penalization for adversarial domain adaptation.
\newblock In {\em International conference on machine learning}, pages 1081--1090. PMLR, 2019.

\bibitem[\protect\citeauthoryear{Chen \bgroup \em et al.\egroup }{2023}]{chen-et-al:scheme}
Hao Chen, Jindong Wang, Ankit Shah, Ran Tao, Hongxin Wei, Xing Xie, Masashi Sugiyama, and Bhiksha Raj.
\newblock Understanding and mitigating the label noise in pre-training on downstream tasks, 2023.

\bibitem[\protect\citeauthoryear{Croce \bgroup \em et al.\egroup }{2023}]{croce-et-al:scheme}
Francesco Croce, Sylvestre-Alvise Rebuffi, Evan Shelhamer, and Sven Gowal.
\newblock Seasoning model soups for robustness to adversarial and natural distribution shifts.
\newblock In {\em Proceedings of the IEEE/CVF Conference on Computer Vision and Pattern Recognition}, pages 12313--12323, 2023.

\bibitem[\protect\citeauthoryear{Devlin \bgroup \em et al.\egroup }{2018}]{devlin2018bert}
Jacob Devlin, Ming-Wei Chang, Kenton Lee, and Kristina Toutanova.
\newblock Bert: Pre-training of deep bidirectional transformers for language understanding.
\newblock {\em arXiv preprint arXiv:1810.04805}, 2018.

\bibitem[\protect\citeauthoryear{Ding \bgroup \em et al.\egroup }{2022}]{ding-et-al:scheme}
Ning Ding, Yujia Qin, Guang Yang, Fuchao Wei, Zonghan Yang, Yusheng Su, Shengding Hu, Yulin Chen, Chi-Min Chan, Weize Chen, et~al.
\newblock Delta tuning: A comprehensive study of parameter efficient methods for pre-trained language models.
\newblock {\em arXiv preprint arXiv:2203.06904}, 2022.

\bibitem[\protect\citeauthoryear{Gottlob \bgroup \em et al.\egroup }{2002}]{gls:hypertrees}
Georg Gottlob, Nicola Leone, and Francesco Scarcello.
\newblock Hypertree decompositions and tractable queries.
\newblock {\em Journal of Computer and System Sciences}, 64(3):579--627, May 2002.

\bibitem[\protect\citeauthoryear{Gottlob}{1992}]{gottlob:nonmon}
Georg Gottlob.
\newblock Complexity results for nonmonotonic logics.
\newblock {\em Journal of Logic and Computation}, 2(3):397--425, June 1992.

\bibitem[\protect\citeauthoryear{Houlsby \bgroup \em et al.\egroup }{2019}]{houlsby-et-al:scheme}
Neil Houlsby, Andrei Giurgiu, Stanislaw Jastrzebski, Bruna Morrone, Quentin De~Laroussilhe, Andrea Gesmundo, Mona Attariyan, and Sylvain Gelly.
\newblock Parameter-efficient transfer learning for nlp.
\newblock In {\em International Conference on Machine Learning}, pages 2790--2799. PMLR, 2019.

\bibitem[\protect\citeauthoryear{Hu \bgroup \em et al.\egroup }{2021}]{hu-et-al:scheme}
Edward~J Hu, Yelong Shen, Phillip Wallis, Zeyuan Allen-Zhu, Yuanzhi Li, Shean Wang, Lu~Wang, and Weizhu Chen.
\newblock Lora: Low-rank adaptation of large language models.
\newblock {\em arXiv preprint arXiv:2106.09685}, 2021.

\bibitem[\protect\citeauthoryear{{IJCAI Proceedings}}{}]{proceedings}
{IJCAI Proceedings}.
\newblock {IJCAI} camera ready submission.
\newblock \url{https://proceedings.ijcai.org/info}.

\bibitem[\protect\citeauthoryear{Ji \bgroup \em et al.\egroup }{2023}]{ji-et-al:scheme}
Yunjie Ji, Yong Deng, Yan Gong, Yiping Peng, Qiang Niu, Lei Zhang, Baochang Ma, and Xiangang Li.
\newblock Exploring the impact of instruction data scaling on large language models: An empirical study on real-world use cases.
\newblock {\em arXiv preprint arXiv:2303.14742}, 2023.

\bibitem[\protect\citeauthoryear{Kumar \bgroup \em et al.\egroup }{2022}]{kumar-et-al:scheme}
Ananya Kumar, Aditi Raghunathan, Robbie Jones, Tengyu Ma, and Percy Liang.
\newblock Fine-tuning can distort pretrained features and underperform out-of-distribution, 2022.

\bibitem[\protect\citeauthoryear{Levesque}{1984a}]{levesque:functional-foundations}
Hector~J. Levesque.
\newblock Foundations of a functional approach to knowledge representation.
\newblock {\em Artificial Intelligence}, 23(2):155--212, July 1984.

\bibitem[\protect\citeauthoryear{Levesque}{1984b}]{levesque:belief}
Hector~J. Levesque.
\newblock A logic of implicit and explicit belief.
\newblock In {\em Proceedings of the Fourth National Conference on Artificial Intelligence}, pages 198--202, Austin, Texas, August 1984. American Association for Artificial Intelligence.

\bibitem[\protect\citeauthoryear{Li and Liang}{2021}]{li-et-al:scheme}
Xiang~Lisa Li and Percy Liang.
\newblock Prefix-tuning: Optimizing continuous prompts for generation.
\newblock {\em arXiv preprint arXiv:2101.00190}, 2021.

\bibitem[\protect\citeauthoryear{Lin \bgroup \em et al.\egroup }{2021}]{linfednlp:scheme}
Bill~Yuchen Lin, Chaoyang He, Zihang Zeng, Hulin Wang, Yufen Huang, Christophe Dupuy, Rahul Gupta, Mahdi Soltanolkotabi, Xiang Ren, and Salman Avestimehr.
\newblock Fednlp: Benchmarking federated learning methods for natural language processing tasks.
\newblock {\em arXiv preprint arXiv:2104.08815}, 2021.

\bibitem[\protect\citeauthoryear{Liu \bgroup \em et al.\egroup }{2019}]{liu2019roberta}
Yinhan Liu, Myle Ott, Naman Goyal, Jingfei Du, Mandar Joshi, Danqi Chen, Omer Levy, Mike Lewis, Luke Zettlemoyer, and Veselin Stoyanov.
\newblock Roberta: A robustly optimized bert pretraining approach.
\newblock {\em arXiv preprint arXiv:1907.11692}, 2019.

\bibitem[\protect\citeauthoryear{Liu \bgroup \em et al.\egroup }{2022}]{liu2022few}
Haokun Liu, Derek Tam, Mohammed Muqeeth, Jay Mohta, Tenghao Huang, Mohit Bansal, and Colin~A Raffel.
\newblock Few-shot parameter-efficient fine-tuning is better and cheaper than in-context learning.
\newblock {\em Advances in Neural Information Processing Systems}, 35:1950--1965, 2022.

\bibitem[\protect\citeauthoryear{Luo \bgroup \em et al.\egroup }{2021}]{luo-et-al:scheme}
Mi~Luo, Fei Chen, Dapeng Hu, Yifan Zhang, Jian Liang, and Jiashi Feng.
\newblock No fear of heterogeneity: Classifier calibration for federated learning with non-iid data.
\newblock {\em Advances in Neural Information Processing Systems}, 34:5972--5984, 2021.

\bibitem[\protect\citeauthoryear{McAuley and Leskovec}{2013}]{amazon}
Julian McAuley and Jure Leskovec.
\newblock Hidden factors and hidden topics: understanding rating dimensions with review text.
\newblock In {\em Proceedings of the 7th ACM conference on Recommender systems}, pages 165--172, 2013.

\bibitem[\protect\citeauthoryear{Muandet \bgroup \em et al.\egroup }{2013}]{Muandet-et-al:scheme}
Krikamol Muandet, David Balduzzi, and Bernhard Sch{\"o}lkopf.
\newblock Domain generalization via invariant feature representation.
\newblock In {\em International conference on machine learning}, pages 10--18. PMLR, 2013.

\bibitem[\protect\citeauthoryear{Nebel}{2000}]{nebel:jair-2000}
Bernhard Nebel.
\newblock On the compilability and expressive power of propositional planning formalisms.
\newblock {\em Journal of Artificial Intelligence Research}, 12:271--315, 2000.

\bibitem[\protect\citeauthoryear{Potts \bgroup \em et al.\egroup }{2020}]{potts2020dynasent}
Christopher Potts, Zhengxuan Wu, Atticus Geiger, and Douwe Kiela.
\newblock Dynasent: A dynamic benchmark for sentiment analysis.
\newblock {\em arXiv preprint arXiv:2012.15349}, 2020.

\bibitem[\protect\citeauthoryear{Qu \bgroup \em et al.\egroup }{2022}]{qu2022rethinking}
Liangqiong Qu, Yuyin Zhou, Paul~Pu Liang, Yingda Xia, Feifei Wang, Ehsan Adeli, Li~Fei-Fei, and Daniel Rubin.
\newblock Rethinking architecture design for tackling data heterogeneity in federated learning.
\newblock In {\em Proceedings of the IEEE/CVF Conference on Computer Vision and Pattern Recognition}, pages 10061--10071, 2022.

\bibitem[\protect\citeauthoryear{Raffel \bgroup \em et al.\egroup }{2020}]{t5raffel2020exploring}
Colin Raffel, Noam Shazeer, Adam Roberts, Katherine Lee, Sharan Narang, Michael Matena, Yanqi Zhou, Wei Li, and Peter~J Liu.
\newblock Exploring the limits of transfer learning with a unified text-to-text transformer.
\newblock {\em The Journal of Machine Learning Research}, 21(1):5485--5551, 2020.

\bibitem[\protect\citeauthoryear{Rebuffi \bgroup \em et al.\egroup }{2021}]{rebuffi-et-al:scheme}
Sylvestre-Alvise Rebuffi, Sven Gowal, Dan~Andrei Calian, Florian Stimberg, Olivia Wiles, and Timothy~A Mann.
\newblock Data augmentation can improve robustness.
\newblock {\em Advances in Neural Information Processing Systems}, 34:29935--29948, 2021.

\bibitem[\protect\citeauthoryear{Saif \bgroup \em et al.\egroup }{2012}]{semeval}
Hassan Saif, Yulan He, and Harith Alani.
\newblock Semantic sentiment analysis of twitter.
\newblock In {\em The Semantic Web--ISWC 2012: 11th International Semantic Web Conference, Boston, MA, USA, November 11-15, 2012, Proceedings, Part I 11}, pages 508--524. Springer, 2012.

\bibitem[\protect\citeauthoryear{Shi \bgroup \em et al.\egroup }{2022}]{shi-et-al:scheme}
Yujun Shi, Jian Liang, Wenqing Zhang, Vincent~YF Tan, and Song Bai.
\newblock Towards understanding and mitigating dimensional collapse in heterogeneous federated learning.
\newblock {\em arXiv preprint arXiv:2210.00226}, 2022.

\bibitem[\protect\citeauthoryear{Socher \bgroup \em et al.\egroup }{2013}]{sst}
Richard Socher, Alex Perelygin, Jean Wu, Jason Chuang, Christopher~D Manning, Andrew~Y Ng, and Christopher Potts.
\newblock Recursive deep models for semantic compositionality over a sentiment treebank.
\newblock In {\em Proceedings of the 2013 conference on empirical methods in natural language processing}, pages 1631--1642, 2013.

\bibitem[\protect\citeauthoryear{Touvron \bgroup \em et al.\egroup }{2023}]{touvron2023llama}
Hugo Touvron, Thibaut Lavril, Gautier Izacard, Xavier Martinet, Marie-Anne Lachaux, Timoth{\'e}e Lacroix, Baptiste Rozi{\`e}re, Naman Goyal, Eric Hambro, Faisal Azhar, et~al.
\newblock Llama: Open and efficient foundation language models.
\newblock {\em arXiv preprint arXiv:2302.13971}, 2023.

\bibitem[\protect\citeauthoryear{Wang \bgroup \em et al.\egroup }{2022}]{wang2022does}
Lirui Wang, Kaiqing Zhang, Yunzhu Li, Yonglong Tian, and Russ Tedrake.
\newblock Does learning from decentralized non-iid unlabeled data benefit from self supervision?
\newblock In {\em The Eleventh International Conference on Learning Representations}, 2022.

\bibitem[\protect\citeauthoryear{Wang \bgroup \em et al.\egroup }{2023}]{wang-et-al:scheme}
Jindong Wang, Xixu Hu, Wenxin Hou, Hao Chen, Runkai Zheng, Yidong Wang, Linyi Yang, Haojun Huang, Wei Ye, Xiubo Geng, et~al.
\newblock On the robustness of chatgpt: An adversarial and out-of-distribution perspective.
\newblock {\em arXiv preprint arXiv:2302.12095}, 2023.

\bibitem[\protect\citeauthoryear{Wortsman \bgroup \em et al.\egroup }{2022}]{wortsman-et-al:scheme}
Mitchell Wortsman, Gabriel Ilharco, Jong~Wook Kim, Mike Li, Simon Kornblith, Rebecca Roelofs, Raphael~Gontijo Lopes, Hannaneh Hajishirzi, Ali Farhadi, Hongseok Namkoong, et~al.
\newblock Robust fine-tuning of zero-shot models.
\newblock In {\em Proceedings of the IEEE/CVF Conference on Computer Vision and Pattern Recognition}, pages 7959--7971, 2022.

\bibitem[\protect\citeauthoryear{Ye \bgroup \em et al.\egroup }{2022}]{ye-et-al:scheme}
Nanyang Ye, Kaican Li, Haoyue Bai, Runpeng Yu, Lanqing Hong, Fengwei Zhou, Zhenguo Li, and Jun Zhu.
\newblock Ood-bench: Quantifying and understanding two dimensions of out-of-distribution generalization.
\newblock In {\em Proceedings of the IEEE/CVF Conference on Computer Vision and Pattern Recognition}, pages 7947--7958, 2022.

\bibitem[\protect\citeauthoryear{Yuan \bgroup \em et al.\egroup }{2023}]{yuan2023revisiting}
Lifan Yuan, Yangyi Chen, Ganqu Cui, Hongcheng Gao, Fangyuan Zou, Xingyi Cheng, Heng Ji, Zhiyuan Liu, and Maosong Sun.
\newblock Revisiting out-of-distribution robustness in nlp: Benchmark, analysis, and llms evaluations.
\newblock {\em arXiv preprint arXiv:2306.04618}, 2023.

\bibitem[\protect\citeauthoryear{Zaken \bgroup \em et al.\egroup }{2021}]{zaken-et-al:scheme}
Elad~Ben Zaken, Shauli Ravfogel, and Yoav Goldberg.
\newblock Bitfit: Simple parameter-efficient fine-tuning for transformer-based masked language-models.
\newblock {\em arXiv preprint arXiv:2106.10199}, 2021.

\bibitem[\protect\citeauthoryear{Zeng \bgroup \em et al.\egroup }{2023}]{zeng2023fedlab}
Dun Zeng, Siqi Liang, Xiangjing Hu, Hui Wang, and Zenglin Xu.
\newblock Fedlab: A flexible federated learning framework.
\newblock {\em Journal of Machine Learning Research}, 24(100):1--7, 2023.

\bibitem[\protect\citeauthoryear{Zhang \bgroup \em et al.\egroup }{2023}]{zhang2023fedpetuning}
Zhuo Zhang, Yuanhang Yang, Yong Dai, Qifan Wang, Yue Yu, Lizhen Qu, and Zenglin Xu.
\newblock Fedpetuning: When federated learning meets the parameter-efficient tuning methods of pre-trained language models.
\newblock In {\em Annual Meeting of the Association of Computational Linguistics 2023}, pages 9963--9977. Association for Computational Linguistics (ACL), 2023.

\bibitem[\protect\citeauthoryear{Zhou \bgroup \em et al.\egroup }{2022}]{zhou-et-al:scheme}
Kaiyang Zhou, Ziwei Liu, Yu~Qiao, Tao Xiang, and Chen~Change Loy.
\newblock Domain generalization: A survey.
\newblock {\em IEEE Transactions on Pattern Analysis and Machine Intelligence}, 2022.

\bibitem[\protect\citeauthoryear{Zhou \bgroup \em et al.\egroup }{2023}]{zhou-et-al:scheme1}
Chunting Zhou, Pengfei Liu, Puxin Xu, Srini Iyer, Jiao Sun, Yuning Mao, Xuezhe Ma, Avia Efrat, Ping Yu, Lili Yu, et~al.
\newblock Lima: Less is more for alignment.
\newblock {\em arXiv preprint arXiv:2305.11206}, 2023.

\end{thebibliography}

\end{document}